\pdfoutput=1
\documentclass[11pt]{article}
\usepackage{graphicx} 
\usepackage{float} 
\usepackage{subfigure} 
\usepackage{graphicx}
\usepackage{enumitem}
\usepackage{verbatim}
\usepackage[]{emnlp2021}
\usepackage{booktabs}
\usepackage{multirow}
\usepackage{makecell}
\usepackage{times}
\usepackage{latexsym}

\newcommand*{\affaddr}[1]{#1}
\newcommand*{\affmark}[1][*]{\textsuperscript{#1}}

\newcommand{\LikertDistributions}{
\begin{figure}[t!]
\centering 
\includegraphics[scale=0.38]{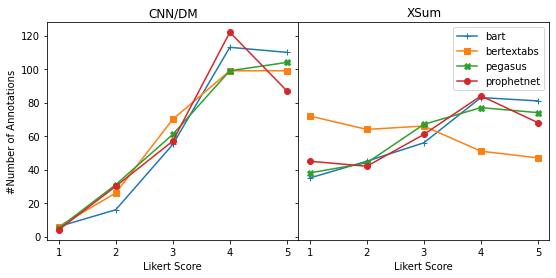}
\caption{Score distribution of LS with a 5-point scale across CNN/DM and XSum. Each data point shows the number of times a score was assigned to each system.} 
\label{Fig.main1} 
\end{figure}
\setlength{\belowcaptionskip}{-0.4cm}
}

\newcommand{\LikertDistributionsb}{
\begin{figure}[t!]
\centering 
\includegraphics[scale=0.38]{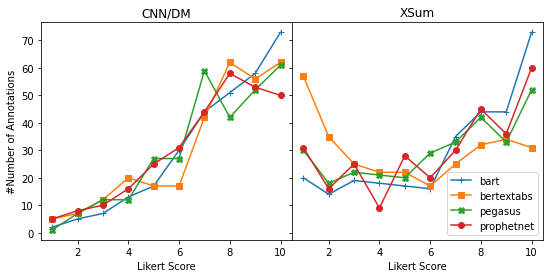}
\caption{Score distribution of $LS_{10}$ across CNN/DM and XSum. Each data point shows the number of times a score was assigned to each system.} 
\label{Fig.main2} 
\end{figure}
}

\newcommand{\BwsInstructions}{
\begin{figure*}[h]
  \centering
  \includegraphics[width=0.9\textwidth]{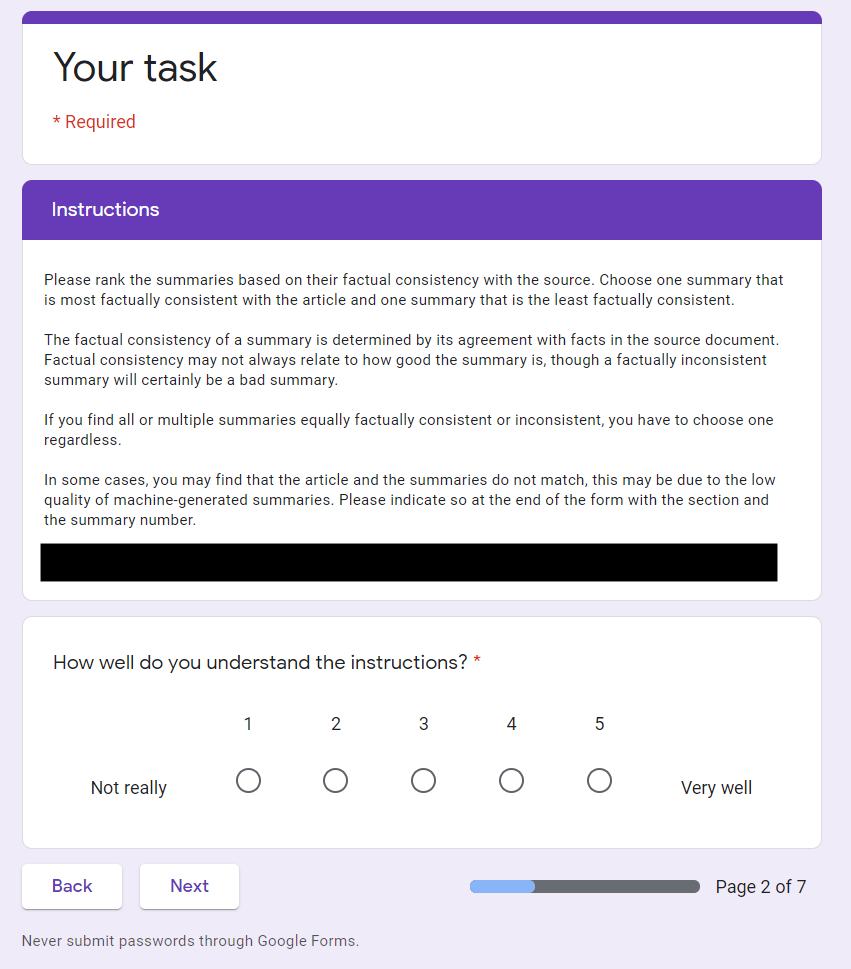}
  \caption{Screenshot of the instruction page for BWS annotation.}
  \label{fig.3}
\end{figure*}
}

\newcommand{\BwsEvaluation}{
\begin{figure*}[h]
  \centering
  \includegraphics[width=0.9\textwidth]{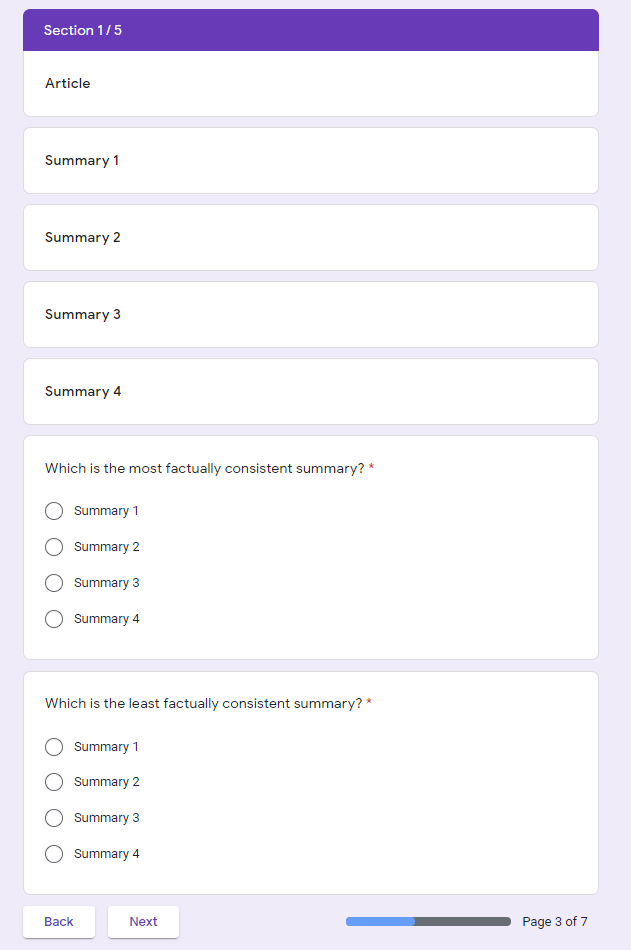}
  \caption{Screenshot of the evaluation page for BWS annotation.}
  \label{fig.4}
\end{figure*}
}

\newcommand{\LikertInstructions}{
\begin{figure*}[h]
  \centering
  \includegraphics[width=0.9\textwidth]{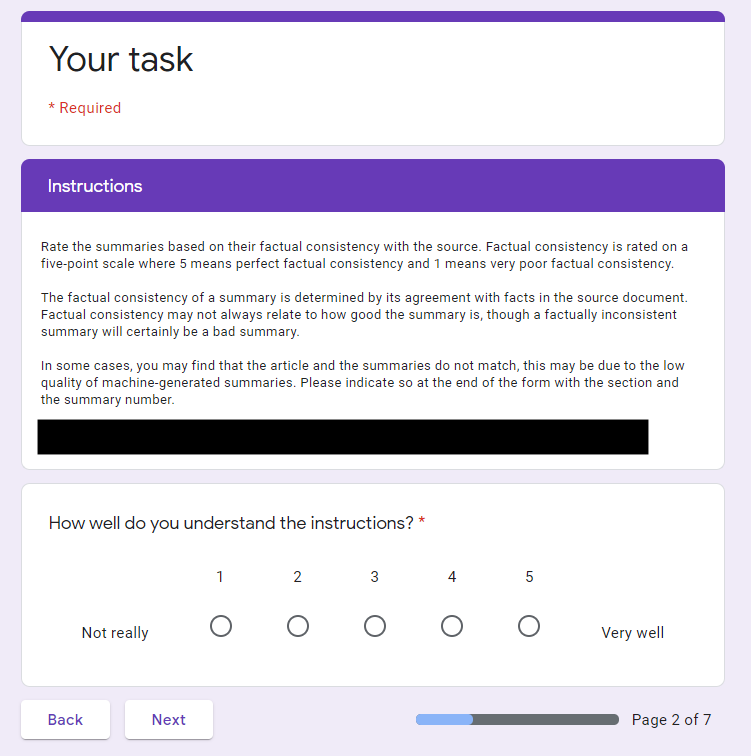}
  \caption{Screenshot of the instruction page we used for Likert Scale annotation.}
  \label{fig.5}
\end{figure*}
}

\newcommand{\LikertEvaluation}{
\begin{figure*}[h]
  \centering
  \includegraphics[width=0.9\textwidth]{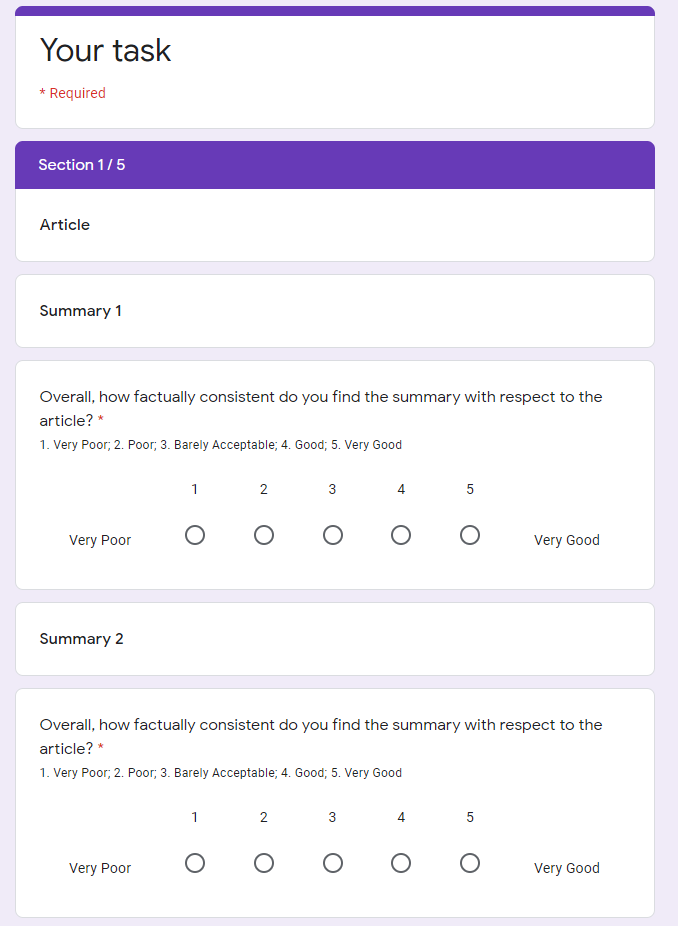}
  \caption{Screenshot of the evaluation page for Likert Scale annotation.}
  \label{fig.6}
\end{figure*}
}

\newcommand{\Sandbox}{
\begin{figure*}[h]
  \centering
  \includegraphics[width=1\textwidth]{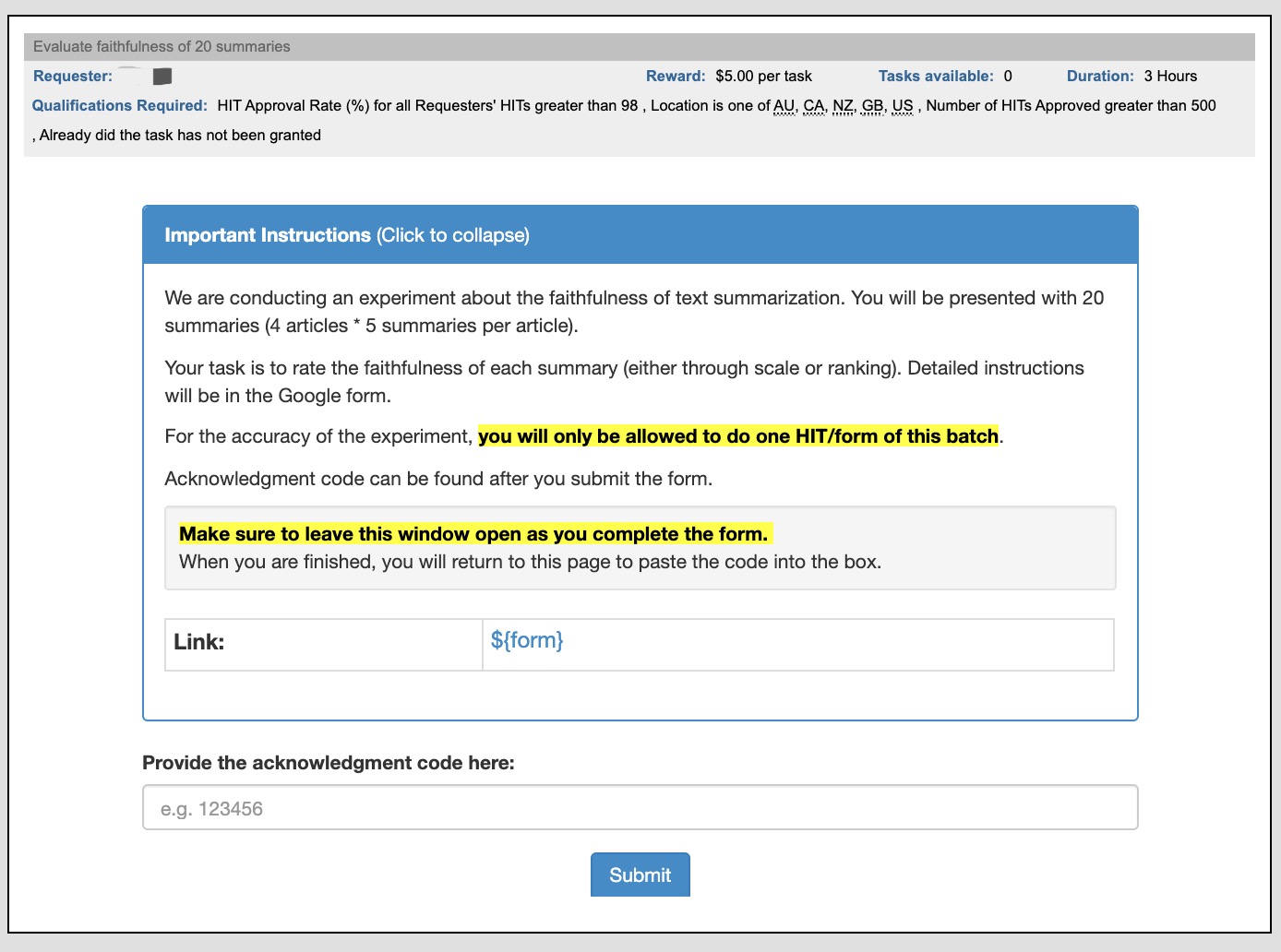}
  \caption{This is how our task will look to Mechanical Turk Workers.}
  \label{fig.sandbox}
\end{figure*}
}

\newcommand{\MLSModelScores}{
\begin{table}[t!]
\centering
\resizebox{\columnwidth}{!}{\begin{tabular}{c | c c |  c c} \Xhline{3\arrayrulewidth}
            &  \multicolumn{2}{c|}{CNN/DM} &  \multicolumn{2}{c}{XSum}  \\ 
            \multirow{-2}{*}{Models}   & LS & $LS_{10}$  & LS & $LS_{10}$ \\ \hline
PEGASUS       &   $3.887^{\color{cyan}2}$ & $7.410^{\color{cyan}3}$    &    $3.350^{\color{red}1}$ & $6.247^{\color{cyan}2}$  \\
ProphetNet      &    $3.860^{\color{blue}4}$  & $7.250^{\color{blue}4}$  &    $3.293^{\color{cyan}3}$ & $6.427^{\color{cyan}2}$  \\
BART          &    $4.017^{\color{red}1}$  & $7.727^{\color{red}1}$ &   $3.433^{\color{cyan}2}$ & $6.937^{\color{red}1}$  \\
 BERTSUM   &    $3.863^{\color{cyan}3}$   & $7.453^{\color{cyan}2}$  &    $2.790^{\color{blue}4}$  & $5.163^{\color{blue}4}$  \\ 
 \Xhline{2\arrayrulewidth}
\end{tabular}}
\caption{Average model rank and average rating scores across LS (5-point scale) and $LS_{10}$ (10-point scale).}
\label{tab:model_scores}
\end{table}
}

\newcommand{\Interchangerate}{
\begin{table}[t!]
\centering
\resizebox{\columnwidth}{!}{\begin{tabular}{c | c c c |  c c c} \Xhline{3\arrayrulewidth}
            &  \multicolumn{3}{c|}{CNN/DM} &  \multicolumn{3}{c}{XSum}  \\ 
            \multirow{0}{0.3cm}{}  & BWS & LS & $LS_{10}$ & BWS & LS & $LS_{10}$ \\ \hline
  Change Rate (\%)     & \textbf{74.71}   & 87.75 &    96.00  &   \textbf{70.25} & 92.25   &  96.25 \\
  Scale Overlap &  -  & 0.67  &  0.61    &  - & \textbf{0.88}  &  \textbf{0.82} \\
 \Xhline{2\arrayrulewidth}
\end{tabular}}
\caption{Change Rate, or percentage of summaries given different ranks or ratings by different annotators (lower is better). Scale Overlap, or average overlap of the range of rating scores between annotators (higher is better).}
\label{tab:change_rate}
\end{table}
}

\newcommand{\RougeScores}{
\begin{table}[t!]
\centering
\resizebox{\columnwidth}{!}{\begin{tabular}{c | c c c |  c c c} \Xhline{3\arrayrulewidth}
            &  \multicolumn{3}{c|}{CNN/DM} &  \multicolumn{3}{c}{XSum}  \\ 
           \multirow{-3}{*}{Models}  & R-1 & R-2 & R-L & R-1 & R-2 & R-L \\ \hline
            PEGASUS       & $44.19^{\color{red}1}$ &  $21.45^{\color{red}1}$ & $41.08^{\color{red}1}$  & $46.84^{\color{red}1}$   & $24.52^{\color{red}1}$   & $39.10^{\color{red}1}$  \\ 
            ProphetNet    &  $42.45^{\color{cyan}3}$   & $19.90^{\color{cyan}3}$   & $39.31^{\color{cyan}3}$  & $43.23^{\color{cyan}3}$  & $19.96^{\color{cyan}3}$   & $35.16^{\color{cyan}3}$  \\
        BART          & $44.07^{\color{cyan}2}$   & $21.13^{\color{cyan}2}$ & $40.89^{\color{cyan}2}$  & $44.15^{\color{cyan}2}$   & $21.28^{\color{cyan}2}$   & $35.94^{\color{cyan}2}$  \\
        BERTSUM & $41.82^{\color{blue}4}$   & $19.39^{\color{blue}4}$    & $38.67^{\color{blue}4}$ & $38.21^{\color{blue}4}$   & $16.11^{\color{blue}4}$    & $30.83^{\color{blue}4}$    \\     \Xhline{2\arrayrulewidth}
\end{tabular}}
\caption{ROUGE-1/2/L scores for model reproduction on CNN/DM and XSum datasets. We apply models directly when they are already fine-tuned and otherwise re-trained them. Pegasus and BART generally obtain the highest ROUGE scores, with ProphetNet comparable in both cases and BERTSUM notably worse on XSum.}
\label{tab:rouge_scores}
\end{table}
}

\newcommand{\ReliabilityScores}{
\begin{table}[t!]
\centering
\resizebox{\columnwidth}{!}{\begin{tabular}{c | c c |  c c} \Xhline{3\arrayrulewidth}
            &  \multicolumn{2}{c|}{CNN/DM} &  \multicolumn{2}{c}{XSum}  \\ 
            
          \multirow{-3}{*}{\centering Scale}  & $\alpha$ & SHR & $\alpha$ & SHR \\ 
                  \hline
                      \multicolumn{5}{c}{\it Protocols} \\
          \hline
          LS & $4.43$ &  $45.61$  & $22.02$  &  $\textbf{92.77}$ \\
          BWS & $\textbf{15.82}$ & $\textbf{87.65}$ & $\textbf{24.77}$ & $90.31$  \\
          \hline
                      \multicolumn{5}{c}{\it Ours} \\
          \hline
                              $LS_{10}$ & $12.87$ & $51.36$  & $29.51$ & $\textbf{94.85}$ \\ $BWS_{value}$ & $\textbf{29.31}$ & $\textbf{92.48}$  & $\textbf{30.62}$ & $92.98$ \\
          \Xhline{2\arrayrulewidth}
\end{tabular}}
\caption{Instance and system-level reliability computed by Krippendorff's alpha ($\alpha$) and split-half reliability (SHR) on the CNN/DM and XSum datasets.}
\label{tab:reliability_scores}
\end{table}
}

\usepackage[T1]{fontenc}

\usepackage[utf8]{inputenc}

\usepackage{microtype}

\title{Investigating Crowdsourcing Protocols for Evaluating \\ the Factual Consistency of Summaries}
\begin{document}

\author{
\textbf{Xiangru Tang}\affmark[$\dagger$]  \quad
 \textbf{Alexander R. Fabbri}\affmark[$\dagger$]  \quad
 \textbf{Haoran Li }\affmark[$\ddagger$]
    \quad  
   \textbf{Ziming Mao}\affmark[$\dagger$]  \quad \textbf{Griffin Adams}\affmark[$\P$]  \quad \textbf{Borui Wang}\affmark[$\dagger$]\\
   \quad \textbf{Asli Celikyilmaz}\affmark[$\ddagger$] 
  \quad \textbf{Yashar Mehdad}\affmark[$\ddagger$] \quad \textbf{Dragomir Radev}\affmark[$\dagger$] \\
\affaddr{\affmark[$\dagger$] Yale University} 
\affaddr{\affmark[$\P$] Columbia University} 
\affaddr{\affmark[$\ddagger$] Facebook AI} \\
\texttt{\{xiangru.tang, alexander.fabbri, ziming.mao,}\\ \texttt{borui.wang, dragomir.radev\}@yale.edu} \\
\texttt{\{aimeeli, aslic, mehdad\}@fb.com} 
}
\maketitle

\begin{abstract}
Current pre-trained models applied for summarization are prone to factual inconsistencies that misrepresent the source text. 
Evaluating the factual consistency of summaries is thus necessary to develop better models. 
However, the human evaluation setup for evaluating factual consistency has not been standardized.
To determine the factors that affect the reliability of the human evaluation, we crowdsource evaluations for factual consistency across state-of-the-art models on two news summarization datasets using the rating-based Likert Scale and ranking-based Best-Worst Scaling. 
Our analysis reveals that the ranking-based Best-Worst Scaling offers a more reliable measure of summary quality across datasets and that the reliability of Likert ratings highly depends on the target dataset and the evaluation design.
To improve crowdsourcing reliability,  we extend the scale of the Likert rating and present a scoring algorithm for Best-Worst Scaling that we call \textit{value learning}. 
Our crowdsourcing guidelines will be publicly available to facilitate future work on factual consistency in summarization.
\end{abstract}
\section{Introduction}
Pre-trained language models have achieved promising results in abstractive text summarization~\cite{edunov-etal-2019-pre, NEURIPS2019_c20bb2d9, song2019mass, zhang-etal-2019-hibert, zhang2020pegasus}. 
A serious limitation of these models, however, is their tendency to produce text that is factually inconsistent with the input.  
Thus, evaluating the factual consistency of the generated summaries with respect to the source is an important task~\cite{falke-etal-2019-ranking, cao-etal-2020-factual,gabriel2020go, durmus-etal-2020-feqa,huang2021factual,pagnoni2021understanding}.
\par
Recently, metrics have been proposed for evaluating factual consistency, including applying natural language inference~\cite{falke-etal-2019-ranking,kryscinski-etal-2020-evaluating} and question-answering models~\cite{eyal-etal-2019-question,scialom-etal-2019-answers,durmus-etal-2020-feqa,wang2020asking}. 
However, current metrics still do not correlate highly with human judgments on factual consistency~\cite{koto2020ffci,pagnoni2021understanding}.
%
To overcome the inherent limitation of automatic metrics, researchers typically crowdsource human evaluations using platforms such as Amazon’s Mechanical Turk (MTurk)~\cite{gillick-liu-2010-non,sabou2012crowdsourcing,lloret2013analyzing}. 
%
%
However, papers often differ in their preferred evaluation protocols \cite{louis-nenkova-2013-automatically, hardy-etal-2019-highres}.  
These differences in the evaluation task design affect the quality of the resulting human judgments and system comparisons \cite{Santhanam2019TowardsBE}. 
%

Two of the primary paradigms of crowdsourced evaluations are ranking-based and rating-based.
Best-Worst Scaling ~\cite{louviere1991best} is a ranking-based method by which the annotator selects the best and worst example out of a set of examples. 
Prior research has claimed that Best-Worst Scaling produces higher-quality evaluations than rating scales such as the Likert Scale for tasks such as sentiment analysis ~\cite{kiritchenko-mohammad-2017-best}.
In the context of summarization, \citet{steen2021evaluate} find that, compared to the Likert Scale, ranking-based protocols are more reliable for measuring summary coherence but less so for repetition. 
However, previous studies have not analyzed annotation reliability in the context of factual consistency for summarization. 
\par 
Our contributions are the following: 1) We are, to the best of our knowledge, the first to study the reliability of human evaluation for summarization factual consistency.  
2) We study rating and ranking-based protocols across two summarization datasets and four state-of-the-art abstractive models. 
We determine  the  factors affecting human evaluation reliability and present a novel ranking-based protocol with the highest reliability.
3) We will release our evaluation guidelines and annotations to promote future work on factual consistency evaluation.

\section{Study Design}
\vspace{-0.1cm}
Each study consists of 100 input documents randomly sampled from each dataset, and four associated model-generated summaries.
\RougeScores
\vspace{-0.1cm}

\subsection{Datasets and Models}
\textbf{Datasets:}
We conduct our study on two benchmark summarization datasets. 
CNN/DailyMail \cite{hermann2015teaching,nallapati2016abstractive} consists of 311,672 pairs of online articles and bullet-point summaries, typically three sentences.  
XSum~\cite{narayan-etal-2018-dont} consists of 227K online articles and single-sentence summaries. 
\par \noindent
\textbf{Models:}
The following abstractive summarization models are chosen due to their strong cross-dataset performance:
\textbf{BART}~\cite{lewis-etal-2020-bart}, a denoising autoencoder for pretraining sequence to sequence and natural language understanding tasks; \textbf{ProphetNet}~\cite{qi-etal-2020-prophetnet}, a pre-trained encoder-decoder model that performs n-gram language modeling;
\textbf{PEGASUS}~\cite{zhang2020pegasus}, a model pre-trained with a summarization-specific objective function; and 
\textbf{BERTSUM}~\cite{liu-lapata-2019-text}, a two-stage fine-tuning approach. Table \ref{tab:rouge_scores} shows the models' ROUGE scores \cite{lin-2004-rouge}.
%
%
\subsection{Reliability}
We follow \citet{steen2021evaluate} and report Krippendorff's alpha and Split-Half Reliability as measures of the reliability of crowdsourced annotations.
\textbf{Krippendorff's alpha ($\alpha$)} is a reliability coefficient developed to measure the agreement among multiple annotators~\cite{krippendorff2011computing}.
This measures instance-level reliability, especially how reliable judgments are over individual summary instances. 
For system-level rankings, to measure the reliability of the rankings of summarization models, we compute \textbf{Split-Half Reliability (SHR)}. 
To compute SHR, annotations are split into two independent groups, and Pearson correlations are calculated between the groups. 
\par
We follow a similar block-design described in \citet{steen2021evaluate}.  
We note that we include the input document as the context of the summaries as opposed to the coherence and repetition dimensions studied in that work, which do not require reading the input article.
We divided our corpus into 20 blocks of 5 documents. We include all 4 generated summaries for each document in the same block, resulting in 5 × 4 = 20 summaries per block. 
We require 3 annotators per block as in \citet{steen2021evaluate}, and each annotator is limited to annotating at most two blocks total across all tasks. 
A further study of the effect of the number of annotators or block design is left for future work.
Crowdsourcing is done via MTurk. 
\subsection{Protocols}
The \textbf{Likert Scale (LS)}  is a common rating-based evaluation protocol \cite{DBLP:conf/ecir/AsgharPHJM18}. 
\MLSModelScores
Likert Scales applied to summarization typically range from 1-5 \cite{steen2021evaluate}.  
\textbf{Best-Worst Scaling (BWS)} is a type of ranking-oriented evaluation that requires annotators to specify only the best and the worst example in a set of summaries \cite{hollis2018best, kiritchenko-mohammad-2017-best}. 
For BWS, the annotator labels the most factually consistent summary and the least factually consistent summary.
Another type of ranking-based protocol is pairwise comparison, where each example is compared to every other example.
However, this protocol is very expensive; given $N$ items to annotate, $N^2$ total annotations must be collected as opposed to BWS which requires a constant factor of $N$ total annotators. 
Due to this exorbitant cost as any reasonable scale, we restrict our study of ranking-based protocols to BWS, and we refer the reader to \citet{kiritchenko-mohammad-2017-best} for an in-depth discussion of the cost comparison for the task of sentiment analysis. 
\subsection{Research Questions}
We study three three main research questions (RQ):
\paragraph{RQ1: Ranking (BWS) vs. LS?} 
We aim to determine the more reliable evaluation protocol. \vspace{-0.2cm}
\paragraph{RQ2: What affects reliability?}
We aim to determine the factors that affect the reliability of the human evaluation. 
\vspace{-0.2cm}
\paragraph{RQ3: What are the protocols' limitations and how to improve them?} Based on the analysis, we propose two protocols to improve the reliability.
\ReliabilityScores
\section{Analysis}
We show the average ratings across \texttt{LS} scales, including a modified LS scale we will later introduce, in Table \ref{tab:model_scores}. 
%
%
Despite the consistently higher ROUGE scores, Pegasus was not always ranked highest, which aligns with previous work suggesting that ROUGE score does not correlate with factual consistency \cite{durmus-etal-2020-feqa}.  
The primary results for reliability evaluation are found in Table \ref{tab:reliability_scores}.
\vspace{-0.15cm}
\paragraph{RQ1: BWS outperforms LS on CNN/DM.}
We see on the left-hand side of the first two rows of Table \ref{tab:reliability_scores} that \texttt{BWS} outperforms \texttt{LS} by a large margin on both instance-level ($\alpha$) and system-level (SHR) reliability. 
As seen in the distribution of the \texttt{LS} ratings in Figures \ref{Fig.main1}, many models are rated as factually consistent with scores of 4 or 5. 
This coincides with previous investigations on CNN/DM which conclude that recent summarization systems produce fluent texts with relatively few factual errors \cite{fabbri2021summeval}.
%
%
We hypothesize that the greater reliability of BWS on CNN/DM data may result from the ranking task forcing the annotator to choose the best summary and distinguish these close summaries rather than allowing e.g. the annotator to give both a score of 5. 
%
%
This result suggests that \texttt{BWS} is preferable in cases where the summaries analyzed have similar factual consistency, such as CNN/DM. 

Though agreement on individual summaries ($\alpha$) is relatively low for all annotation methods, these numbers are comparable to those obtained in \cite{steen2021evaluate}. 
Furthermore, we look at the relative difference between ($\alpha$) of BWS and LS, and we find that studies still arrive at consistent system scores as demonstrated by the SHR. 
This reflects similar observations made by \citet{gillick-liu-2010-non}.
System-level ranks such as SHR, are also more important for evaluation purposes as the goal is generally to rank models to determine the best performing (or most factually consistent) system as opposed to examining individual examples as Krippendorff’s alpha measures. 
\LikertDistributions
\vspace{-0.15cm}

\paragraph{RQ2: Dataset Characteristics Affect Reliability.}
We extend our experiments to the XSum dataset to see whether the reliability of the protocols changes as the characteristics of the dataset change. 
XSum-trained models are known to suffer from factual inconsistencies because of the high compression ratio and high level of abstraction of the reference summaries \cite{maynez2020faithfulness}.
As seen on the right-hand side of the first two rows of Table \ref{tab:reliability_scores}, \texttt{BWS} and \texttt{LS} both perform well, with \texttt{LS} slightly outperforming \texttt{BWS} according to SHR. 
As seen in Figure \ref{Fig.main1}, the model scores are more spread out along the scale. 
This coincides with the large range of ROUGE scores and larger differences between models, as seen in Table \ref{tab:rouge_scores}, which likely explains why annotators can differentiate the model outputs better. 
Thus, we believe that \texttt{LS} is a viable option when the corpus contains a diverse quality of summaries, like XSum.
\vspace{-0.15cm}
\LikertDistributionsb
\paragraph{RQ3: Improvements and Current Limitations.}
We propose two modified protocols to improve reliability and then study the presence of common limitations for evaluation protocols. 
Prior work has noted \textbf{the effect of scale granularity} \cite{kiritchenko-mohammad-2017-best}, so for \texttt{LS}, we extend the scale from for 5 to 10 and call it \texttt{LS-10}.
Table \ref{tab:reliability_scores} shows that that \texttt{LS-10} is more reliable than \texttt{LS}
A finer-grained scale may capture more nuanced differences in data points with more choices. 
Scores tend to move towards the extremes when we use a finer-grained scale (10 vs 5), as seen in the difference in distributions in Figures~\ref{Fig.main1} and~\ref{Fig.main2}. 
Thus, for \texttt{LS-10}, a larger range and being less biased towards a specific region, promoting better reliability.
%
Previous work suggests that Best-Worst Scaling fails to yield an unbiased estimate of the true quality value \cite{hollis2018scoring}. 
Thus, for \texttt{BWS}, we incorporate information about the quality of competing examples or \textit{value learning} into a $BWS_{value}$ protocol. 
The annotator is asked to give a score (3-point scale) for the difference between the best and the worst summary. 
The final ranking uses a weighted sum. 
The results at the bottom of Table \ref{tab:reliability_scores} also confirm the effectiveness of this protocol.
%
%
%

\par
To verify the limitations of evaluation protocols noted by \citet{kiritchenko-mohammad-2017-best}, we conduct the following studies. 
We first analyze \textbf{(a) the inconsistencies in annotations by different annotators}, measured by the percentage of summaries that receive different ratings or rankings from different annotators, which we call \textbf{change rate}. 
As shown in Table \ref{tab:change_rate}, annotators are more likely to agree on the same ranking in \texttt{BWS} as opposed to the same rating for \texttt{LS}.
We further test \textbf{(b) inconsistencies by the same annotator}, in particular whether annotations done by the same worker are consistent over time. 
We ask workers who have previously annotated XSum and CNN/DM samples to re-do their annotations one week after their initial annotations. 
We notified the workers to re-annotate only one week after they finished, instead of at the beginning, as we do not want to introduce design bias. 
In total, 43 workers redid 860 annotations. For \texttt{LS}, the average change in the rating of the two annotations one week apart by the same worker was 0.92. 
%
\par
Additionally, we examine whether \texttt{LS} suffers from \textbf{(c) scale region bias}, where different annotators are often biased towards different parts of the rating scale. 
For a given block and two annotators, we calculate the rating range given by each annotator. 
We then calculate the overlap length between those two ranges divided by the length of the overall range from both annotators. 
We call this the percentage \textbf{scale overlap} and average over all pairs of annotators and blocks. 
For \texttt{LS}, the percentage scale overlap is (0.67, \textbf{0.88}) for (CNN/DM, XSum), respectively, and (0.61, \textbf{0.82}) for \texttt{LS-10}. 
The difference in scale region bias between \texttt{LS} and \texttt{LS-10} is small, but the bias difference between CNN/DM and XSum is notable.
%
Greater diversity in summary quality as in XSum may force the annotators to expand their use of the scale and mitigate region bias, which may explain why \texttt{LS} is better than \texttt{BWS} on XSum as opposed to CNN/DM. 
Future work may investigate further what exactly constitutes too wide of a scaling range. 
\Interchangerate
\section{Conclusion}
\vspace{-0.1cm}
In this paper, we conduct studies to understand and improve the reliability of ranking and rating-based human evaluations of summarization factual consistency.
We find that Best-Worst Scaling is largely reliable, and the Likert scale also has merits, but the proper scaling and dataset characteristics must be carefully studied to ensure its reliability.
We improve these two protocols based on our findings and believe that our studies advance the understanding of both models and metrics as we aim to facilitate factually consistent text generation.
\section{Ethical Considerations} \label{sec:ethics}
\paragraph{Intellectual Properties and Privacy Rights}
All of the datasets (CNN/DM and XSum) used in our study are publicly available. Regarding privacy rights, the authors of the paper completed IRB human subject protection training for conducting this study. We will release the annotations, but rather than releasing the MTurk ID of the worker, we will completely anonymize this ID.
\paragraph{Compensation for Annotators}
 Workers were compensated \$5 per block, calibrated to equal a  \$15/hour payrate. We first annotated examples in-house to determine the required annotation speed. A summary block usually takes around 20 minutes.
\paragraph{Steps Taken to Avoid Potential Problems}
Annotations were completed in the form of a survey on a Google Form. We provided space for the Turkers to provide feedback. We manually uploaded the data points (articles and summaries) used in this study to avoid any offensive content.

\paragraph{The Number of Examples}
We sampled 100 examples from each dataset that did not contain exactly matching summaries.
Both Likert and BWS follow the same block design, which includes the same number of examples per block. With the exception that the BWS annotation asks for the most and least factually consistent summary and the Likert asks for ratings for each individual summary. Due to space requirements, we included further details, images of the interface, in the supplementary material. We pay the same amount per block of annotations. 

\paragraph{Qualifications of MTurk workers}
We use the following qualifications to recruit in total 350 MTurk workers with good track records: 
HIT approval rate greater than or equal to 98\%, number of HITs approved greater than or equal to 500, and located in one of the following English native-speaking countries: Australia, Canada, New Zealand, United Kingdom, United States.

\bibliography{emnlp2021}

\begin{thebibliography}{38}
\expandafter\ifx\csname natexlab\endcsname\relax\def\natexlab#1{#1}\fi

\bibitem[{Asghar et~al.(2018)Asghar, Poupart, Hoey, Jiang, and
  Mou}]{DBLP:conf/ecir/AsgharPHJM18}
Nabiha Asghar, Pascal Poupart, Jesse Hoey, Xin Jiang, and Lili Mou. 2018.
\newblock \href {https://doi.org/10.1007/978-3-319-76941-7\_12} {Affective
  neural response generation}.
\newblock In \emph{Advances in Information Retrieval - 40th European Conference
  on {IR} Research, {ECIR} 2018, Grenoble, France, March 26-29, 2018,
  Proceedings}, volume 10772 of \emph{Lecture Notes in Computer Science}, pages
  154--166. Springer.

\bibitem[{Cao et~al.(2020)Cao, Dong, Wu, and Cheung}]{cao-etal-2020-factual}
Meng Cao, Yue Dong, Jiapeng Wu, and Jackie Chi~Kit Cheung. 2020.
\newblock \href {https://doi.org/10.18653/v1/2020.emnlp-main.506} {Factual
  error correction for abstractive summarization models}.
\newblock In \emph{Proceedings of the 2020 Conference on Empirical Methods in
  Natural Language Processing (EMNLP)}, pages 6251--6258, Online. Association
  for Computational Linguistics.

\bibitem[{Dong et~al.(2019)Dong, Yang, Wang, Wei, Liu, Wang, Gao, Zhou, and
  Hon}]{NEURIPS2019_c20bb2d9}
Li~Dong, Nan Yang, Wenhui Wang, Furu Wei, Xiaodong Liu, Yu~Wang, Jianfeng Gao,
  Ming Zhou, and Hsiao{-}Wuen Hon. 2019.
\newblock \href
  {https://proceedings.neurips.cc/paper/2019/hash/c20bb2d9a50d5ac1f713f8b34d9aac5a-Abstract.html}
  {Unified language model pre-training for natural language understanding and
  generation}.
\newblock In \emph{Advances in Neural Information Processing Systems 32: Annual
  Conference on Neural Information Processing Systems 2019, NeurIPS 2019,
  December 8-14, 2019, Vancouver, BC, Canada}, pages 13042--13054.

\bibitem[{Durmus et~al.(2020)Durmus, He, and Diab}]{durmus-etal-2020-feqa}
Esin Durmus, He~He, and Mona Diab. 2020.
\newblock \href {https://doi.org/10.18653/v1/2020.acl-main.454} {{FEQA}: A
  question answering evaluation framework for faithfulness assessment in
  abstractive summarization}.
\newblock In \emph{Proceedings of the 58th Annual Meeting of the Association
  for Computational Linguistics}, pages 5055--5070, Online. Association for
  Computational Linguistics.

\bibitem[{Edunov et~al.(2019)Edunov, Baevski, and Auli}]{edunov-etal-2019-pre}
Sergey Edunov, Alexei Baevski, and Michael Auli. 2019.
\newblock \href {https://doi.org/10.18653/v1/N19-1409} {Pre-trained language
  model representations for language generation}.
\newblock In \emph{Proceedings of the 2019 Conference of the North {A}merican
  Chapter of the Association for Computational Linguistics: Human Language
  Technologies, Volume 1 (Long and Short Papers)}, pages 4052--4059,
  Minneapolis, Minnesota. Association for Computational Linguistics.

\bibitem[{Eyal et~al.(2019)Eyal, Baumel, and Elhadad}]{eyal-etal-2019-question}
Matan Eyal, Tal Baumel, and Michael Elhadad. 2019.
\newblock \href {https://doi.org/10.18653/v1/N19-1395} {Question answering as
  an automatic evaluation metric for news article summarization}.
\newblock In \emph{Proceedings of the 2019 Conference of the North {A}merican
  Chapter of the Association for Computational Linguistics: Human Language
  Technologies, Volume 1 (Long and Short Papers)}, pages 3938--3948,
  Minneapolis, Minnesota. Association for Computational Linguistics.

\bibitem[{Fabbri et~al.(2021)Fabbri, Kry{\'s}ci{\'n}ski, McCann, Xiong, Socher,
  and Radev}]{fabbri2021summeval}
Alexander~R Fabbri, Wojciech Kry{\'s}ci{\'n}ski, Bryan McCann, Caiming Xiong,
  Richard Socher, and Dragomir Radev. 2021.
\newblock Summeval: Re-evaluating summarization evaluation.
\newblock \emph{Transactions of the Association for Computational Linguistics},
  9:391--409.

\bibitem[{Falke et~al.(2019)Falke, Ribeiro, Utama, Dagan, and
  Gurevych}]{falke-etal-2019-ranking}
Tobias Falke, Leonardo F.~R. Ribeiro, Prasetya~Ajie Utama, Ido Dagan, and Iryna
  Gurevych. 2019.
\newblock \href {https://doi.org/10.18653/v1/P19-1213} {Ranking generated
  summaries by correctness: An interesting but challenging application for
  natural language inference}.
\newblock In \emph{Proceedings of the 57th Annual Meeting of the Association
  for Computational Linguistics}, pages 2214--2220, Florence, Italy.
  Association for Computational Linguistics.

\bibitem[{Gabriel et~al.(2021)Gabriel, Celikyilmaz, Jha, Choi, and
  Gao}]{gabriel2020go}
Saadia Gabriel, Asli Celikyilmaz, Rahul Jha, Yejin Choi, and Jianfeng Gao.
  2021.
\newblock \href {https://doi.org/10.18653/v1/2021.findings-acl.42} {{GO}
  {FIGURE}: A meta evaluation of factuality in summarization}.
\newblock In \emph{Findings of the Association for Computational Linguistics:
  ACL-IJCNLP 2021}, pages 478--487, Online. Association for Computational
  Linguistics.

\bibitem[{Gillick and Liu(2010)}]{gillick-liu-2010-non}
Dan Gillick and Yang Liu. 2010.
\newblock \href {https://aclanthology.org/W10-0722} {Non-expert evaluation of
  summarization systems is risky}.
\newblock In \emph{Proceedings of the {NAACL} {HLT} 2010 Workshop on Creating
  Speech and Language Data with {A}mazon{'}s Mechanical Turk}, pages 148--151,
  Los Angeles. Association for Computational Linguistics.

\bibitem[{Hardy et~al.(2019)Hardy, Narayan, and
  Vlachos}]{hardy-etal-2019-highres}
Hardy Hardy, Shashi Narayan, and Andreas Vlachos. 2019.
\newblock \href {https://doi.org/10.18653/v1/P19-1330} {{H}igh{RES}:
  Highlight-based reference-less evaluation of summarization}.
\newblock In \emph{Proceedings of the 57th Annual Meeting of the Association
  for Computational Linguistics}, pages 3381--3392, Florence, Italy.
  Association for Computational Linguistics.

\bibitem[{Hermann et~al.(2015)Hermann, Kocisk{\'{y}}, Grefenstette, Espeholt,
  Kay, Suleyman, and Blunsom}]{hermann2015teaching}
Karl~Moritz Hermann, Tom{\'{a}}s Kocisk{\'{y}}, Edward Grefenstette, Lasse
  Espeholt, Will Kay, Mustafa Suleyman, and Phil Blunsom. 2015.
\newblock \href
  {https://proceedings.neurips.cc/paper/2015/hash/afdec7005cc9f14302cd0474fd0f3c96-Abstract.html}
  {Teaching machines to read and comprehend}.
\newblock In \emph{Advances in Neural Information Processing Systems 28: Annual
  Conference on Neural Information Processing Systems 2015, December 7-12,
  2015, Montreal, Quebec, Canada}, pages 1693--1701.

\bibitem[{Hollis(2018)}]{hollis2018scoring}
Geoff Hollis. 2018.
\newblock Scoring best-worst data in unbalanced many-item designs, with
  applications to crowdsourcing semantic judgments.
\newblock \emph{Behavior research methods}, 50(2):711--729.

\bibitem[{Hollis and Westbury(2018)}]{hollis2018best}
Geoff Hollis and Chris Westbury. 2018.
\newblock When is best-worst best? a comparison of best-worst scaling, numeric
  estimation, and rating scales for collection of semantic norms.
\newblock \emph{Behavior research methods}, 50(1):115--133.

\bibitem[{Huang et~al.(2021)Huang, Feng, Feng, and Qin}]{huang2021factual}
Yichong Huang, Xiachong Feng, Xiaocheng Feng, and Bing Qin. 2021.
\newblock \href {http://arxiv.org/abs/2104.14839} {The factual inconsistency
  problem in abstractive text summarization: A survey}.

\bibitem[{Kiritchenko and Mohammad(2017)}]{kiritchenko-mohammad-2017-best}
Svetlana Kiritchenko and Saif Mohammad. 2017.
\newblock \href {https://doi.org/10.18653/v1/P17-2074} {Best-worst scaling more
  reliable than rating scales: A case study on sentiment intensity annotation}.
\newblock In \emph{Proceedings of the 55th Annual Meeting of the Association
  for Computational Linguistics (Volume 2: Short Papers)}, pages 465--470,
  Vancouver, Canada. Association for Computational Linguistics.

\bibitem[{Koto et~al.(2020)Koto, Lau, and Baldwin}]{koto2020ffci}
Fajri Koto, Jey~Han Lau, and Timothy Baldwin. 2020.
\newblock \href {https://arxiv.org/abs/2011.13662} {Ffci: A framework for
  interpretable automatic evaluation of summarization}.
\newblock \emph{ArXiv preprint}, abs/2011.13662.

\bibitem[{Krippendorff(2011)}]{krippendorff2011computing}
Klaus Krippendorff. 2011.
\newblock Computing krippendorff's alpha-reliability.

\bibitem[{Kryscinski et~al.(2020)Kryscinski, McCann, Xiong, and
  Socher}]{kryscinski-etal-2020-evaluating}
Wojciech Kryscinski, Bryan McCann, Caiming Xiong, and Richard Socher. 2020.
\newblock \href {https://doi.org/10.18653/v1/2020.emnlp-main.750} {Evaluating
  the factual consistency of abstractive text summarization}.
\newblock In \emph{Proceedings of the 2020 Conference on Empirical Methods in
  Natural Language Processing (EMNLP)}, pages 9332--9346, Online. Association
  for Computational Linguistics.

\bibitem[{Lewis et~al.(2020)Lewis, Liu, Goyal, Ghazvininejad, Mohamed, Levy,
  Stoyanov, and Zettlemoyer}]{lewis-etal-2020-bart}
Mike Lewis, Yinhan Liu, Naman Goyal, Marjan Ghazvininejad, Abdelrahman Mohamed,
  Omer Levy, Veselin Stoyanov, and Luke Zettlemoyer. 2020.
\newblock \href {https://doi.org/10.18653/v1/2020.acl-main.703} {{BART}:
  Denoising sequence-to-sequence pre-training for natural language generation,
  translation, and comprehension}.
\newblock In \emph{Proceedings of the 58th Annual Meeting of the Association
  for Computational Linguistics}, pages 7871--7880, Online. Association for
  Computational Linguistics.

\bibitem[{Lin(2004)}]{lin-2004-rouge}
Chin-Yew Lin. 2004.
\newblock \href {https://aclanthology.org/W04-1013} {{ROUGE}: A package for
  automatic evaluation of summaries}.
\newblock In \emph{Text Summarization Branches Out}, pages 74--81, Barcelona,
  Spain. Association for Computational Linguistics.

\bibitem[{Liu and Lapata(2019)}]{liu-lapata-2019-text}
Yang Liu and Mirella Lapata. 2019.
\newblock \href {https://doi.org/10.18653/v1/D19-1387} {Text summarization with
  pretrained encoders}.
\newblock In \emph{Proceedings of the 2019 Conference on Empirical Methods in
  Natural Language Processing and the 9th International Joint Conference on
  Natural Language Processing (EMNLP-IJCNLP)}, pages 3730--3740, Hong Kong,
  China. Association for Computational Linguistics.

\bibitem[{Lloret et~al.(2013)Lloret, Plaza, and Aker}]{lloret2013analyzing}
Elena Lloret, Laura Plaza, and Ahmet Aker. 2013.
\newblock Analyzing the capabilities of crowdsourcing services for text
  summarization.
\newblock \emph{Language resources and evaluation}, 47(2):337--369.

\bibitem[{Louis and Nenkova(2013)}]{louis-nenkova-2013-automatically}
Annie Louis and Ani Nenkova. 2013.
\newblock \href {https://doi.org/10.1162/COLI_a_00123} {Automatically assessing
  machine summary content without a gold standard}.
\newblock \emph{Computational Linguistics}, 39(2):267--300.

\bibitem[{Louviere and Woodworth(1991)}]{louviere1991best}
Jordan~J Louviere and George~G Woodworth. 1991.
\newblock Best-worst scaling: A model for the largest difference judgments.
\newblock Technical report, Working paper.

\bibitem[{Maynez et~al.(2020)Maynez, Narayan, Bohnet, and
  McDonald}]{maynez2020faithfulness}
Joshua Maynez, Shashi Narayan, Bernd Bohnet, and Ryan McDonald. 2020.
\newblock \href {https://doi.org/10.18653/v1/2020.acl-main.173} {On
  faithfulness and factuality in abstractive summarization}.
\newblock In \emph{Proceedings of the 58th Annual Meeting of the Association
  for Computational Linguistics}, pages 1906--1919, Online. Association for
  Computational Linguistics.

\bibitem[{Nallapati et~al.(2016)Nallapati, Zhou, dos Santos, Gulcehre, and
  Xiang}]{nallapati2016abstractive}
Ramesh Nallapati, Bowen Zhou, Cicero dos Santos, Caglar Gulcehre, and Bing
  Xiang. 2016.
\newblock \href {https://doi.org/10.18653/v1/K16-1028} {Abstractive text
  summarization using sequence-to-sequence {RNN}s and beyond}.
\newblock In \emph{Proceedings of The 20th {SIGNLL} Conference on Computational
  Natural Language Learning}, pages 280--290, Berlin, Germany. Association for
  Computational Linguistics.

\bibitem[{Narayan et~al.(2018)Narayan, Cohen, and
  Lapata}]{narayan-etal-2018-dont}
Shashi Narayan, Shay~B. Cohen, and Mirella Lapata. 2018.
\newblock \href {https://doi.org/10.18653/v1/D18-1206} {Don{'}t give me the
  details, just the summary! topic-aware convolutional neural networks for
  extreme summarization}.
\newblock In \emph{Proceedings of the 2018 Conference on Empirical Methods in
  Natural Language Processing}, pages 1797--1807, Brussels, Belgium.
  Association for Computational Linguistics.

\bibitem[{Pagnoni et~al.(2021)Pagnoni, Balachandran, and
  Tsvetkov}]{pagnoni2021understanding}
Artidoro Pagnoni, Vidhisha Balachandran, and Yulia Tsvetkov. 2021.
\newblock \href {https://doi.org/10.18653/v1/2021.naacl-main.383}
  {Understanding factuality in abstractive summarization with {FRANK}: A
  benchmark for factuality metrics}.
\newblock In \emph{Proceedings of the 2021 Conference of the North American
  Chapter of the Association for Computational Linguistics: Human Language
  Technologies}, pages 4812--4829, Online. Association for Computational
  Linguistics.

\bibitem[{Qi et~al.(2020)Qi, Yan, Gong, Liu, Duan, Chen, Zhang, and
  Zhou}]{qi-etal-2020-prophetnet}
Weizhen Qi, Yu~Yan, Yeyun Gong, Dayiheng Liu, Nan Duan, Jiusheng Chen, Ruofei
  Zhang, and Ming Zhou. 2020.
\newblock \href {https://doi.org/10.18653/v1/2020.findings-emnlp.217}
  {{P}rophet{N}et: Predicting future n-gram for
  sequence-to-{S}equence{P}re-training}.
\newblock In \emph{Findings of the Association for Computational Linguistics:
  EMNLP 2020}, pages 2401--2410, Online. Association for Computational
  Linguistics.

\bibitem[{Sabou et~al.(2012)Sabou, Bontcheva, and
  Scharl}]{sabou2012crowdsourcing}
Marta Sabou, Kalina Bontcheva, and Arno Scharl. 2012.
\newblock Crowdsourcing research opportunities: lessons from natural language
  processing.
\newblock In \emph{Proceedings of the 12th International Conference on
  Knowledge Management and Knowledge Technologies}, pages 1--8.

\bibitem[{Santhanam and Shaikh(2019)}]{Santhanam2019TowardsBE}
Sashank Santhanam and Samira Shaikh. 2019.
\newblock \href {https://doi.org/10.18653/v1/W19-8610} {Towards best experiment
  design for evaluating dialogue system output}.
\newblock In \emph{Proceedings of the 12th International Conference on Natural
  Language Generation}, pages 88--94, Tokyo, Japan. Association for
  Computational Linguistics.

\bibitem[{Scialom et~al.(2019)Scialom, Lamprier, Piwowarski, and
  Staiano}]{scialom-etal-2019-answers}
Thomas Scialom, Sylvain Lamprier, Benjamin Piwowarski, and Jacopo Staiano.
  2019.
\newblock \href {https://doi.org/10.18653/v1/D19-1320} {Answers unite!
  unsupervised metrics for reinforced summarization models}.
\newblock In \emph{Proceedings of the 2019 Conference on Empirical Methods in
  Natural Language Processing and the 9th International Joint Conference on
  Natural Language Processing (EMNLP-IJCNLP)}, pages 3246--3256, Hong Kong,
  China. Association for Computational Linguistics.

\bibitem[{Song et~al.(2019)Song, Tan, Qin, Lu, and Liu}]{song2019mass}
Kaitao Song, Xu~Tan, Tao Qin, Jianfeng Lu, and Tie{-}Yan Liu. 2019.
\newblock \href {http://proceedings.mlr.press/v97/song19d.html} {{MASS:} masked
  sequence to sequence pre-training for language generation}.
\newblock In \emph{Proceedings of the 36th International Conference on Machine
  Learning, {ICML} 2019, 9-15 June 2019, Long Beach, California, {USA}},
  volume~97 of \emph{Proceedings of Machine Learning Research}, pages
  5926--5936. {PMLR}.

\bibitem[{Steen and Markert(2021)}]{steen2021evaluate}
Julius Steen and Katja Markert. 2021.
\newblock \href {https://aclanthology.org/2021.eacl-main.160} {How to evaluate
  a summarizer: Study design and statistical analysis for manual linguistic
  quality evaluation}.
\newblock In \emph{Proceedings of the 16th Conference of the European Chapter
  of the Association for Computational Linguistics: Main Volume}, pages
  1861--1875, Online. Association for Computational Linguistics.

\bibitem[{Wang et~al.(2020)Wang, Cho, and Lewis}]{wang2020asking}
Alex Wang, Kyunghyun Cho, and Mike Lewis. 2020.
\newblock \href {https://doi.org/10.18653/v1/2020.acl-main.450} {Asking and
  answering questions to evaluate the factual consistency of summaries}.
\newblock In \emph{Proceedings of the 58th Annual Meeting of the Association
  for Computational Linguistics}, pages 5008--5020, Online. Association for
  Computational Linguistics.

\bibitem[{Zhang et~al.(2020)Zhang, Zhao, Saleh, and Liu}]{zhang2020pegasus}
Jingqing Zhang, Yao Zhao, Mohammad Saleh, and Peter~J. Liu. 2020.
\newblock \href {http://proceedings.mlr.press/v119/zhang20ae.html} {{PEGASUS:}
  pre-training with extracted gap-sentences for abstractive summarization}.
\newblock In \emph{Proceedings of the 37th International Conference on Machine
  Learning, {ICML} 2020, 13-18 July 2020, Virtual Event}, volume 119 of
  \emph{Proceedings of Machine Learning Research}, pages 11328--11339. {PMLR}.

\bibitem[{Zhang et~al.(2019)Zhang, Wei, and Zhou}]{zhang-etal-2019-hibert}
Xingxing Zhang, Furu Wei, and Ming Zhou. 2019.
\newblock \href {https://doi.org/10.18653/v1/P19-1499} {{HIBERT}: Document
  level pre-training of hierarchical bidirectional transformers for document
  summarization}.
\newblock In \emph{Proceedings of the 57th Annual Meeting of the Association
  for Computational Linguistics}, pages 5059--5069, Florence, Italy.
  Association for Computational Linguistics.

\end{thebibliography}
\bibliographystyle{acl_natbib}

\newpage
\newpage

\appendix
\section{Appendix}
\label{sec:appendix}
Besides the average model rank and average rating scores across BWS, LS-5, and LS-10 evaluations, we also provide standard deviations in Table \ref{tab:model_scores_std}.

\begin{table*}[htbp]
 \centering
\resizebox{\textwidth}{!}{
\begin{tabular}{c | c c c |  c c c} \Xhline{3\arrayrulewidth}
            &  \multicolumn{3}{c|}{CNN/DM} &  \multicolumn{3}{c}{XSum}  \\ 
Models  & BWS & LS & LS-10 & BWS & LS & LS-10 \\ 

            \hline
PEGASUS    &  $3.230^{\color{cyan}2}/1.150$   &    $3.887^{\color{cyan}2}/1.051$ & $7.410^{\color{cyan}3}/2.160$  &  $3.247^{\color{cyan}3}/0.936$   &    $3.350^{\color{red}1}/1.334$ & $6.247^{\color{cyan}2}/2.978$  \\

ProphetNet    & $3.100^{\color{cyan}3}/1.026$    &    $3.860^{\color{blue}4}/0.992$  & $7.250^{\color{blue}4}/2.252$ & $3.360^{\color{cyan}2}/1.102$    &    $3.293^{\color{cyan}3}/1.359$ & $6.427^{\color{cyan}2}/3.038$  \\

BART          &  $3.593^{\color{red}1}/1.113$   &    $4.017^{\color{red}1}/0.973$  & $7.727^{\color{red}1}/2.090$ & $3.570^{\color{red}1}/1.179$   &    $3.433^{\color{cyan}2}/1.338$ & $6.937^{\color{red}1}/2.889$  \\

 BERTSUM & $3.087^{\color{blue}4}/0.984$    &    $3.863^{\color{cyan}3}/1.037$   & $7.453^{\color{cyan}2}/2.309$ &  $2.827^{\color{blue}4}/0.993$    &    $2.790^{\color{blue}4}/1.390$  & $5.163^{\color{blue}4}/3.202$  \\ 
 \Xhline{2\arrayrulewidth}
\end{tabular}}

\caption{Average model rank, rating, and standard deviation across BWS, LS and $LS_{10}$ evaluations.}
\label{tab:model_scores_std}
\end{table*}

To demonstrate our annotation template and facilitate future research, we show the interface for BWS annotations in Figures \ref{fig.3} and \ref{fig.4} and the interface for Likert annotations in Figures \ref{fig.5} and \ref{fig.6}. We made use of the survey feature in Amazon Mechanical Turk (MTurk) to link to these Google Forms in Figure \ref{fig.sandbox}.  
\BwsInstructions
\BwsEvaluation
\LikertInstructions
\LikertEvaluation
\Sandbox
\newpage

\end{document}